\pdfoutput=1

\documentclass[11pt]{article}

\usepackage[final]{acl}

\usepackage{times}
\usepackage{latexsym}

\usepackage[T1]{fontenc}

\usepackage[utf8]{inputenc}

\usepackage{microtype}

\usepackage{inconsolata}

\usepackage{graphicx}

\usepackage{todonotes}
\usepackage{longtable}
\usepackage{booktabs}
\usepackage{comment}
\excludecomment{thesis}

\title{Tokenization and Morphology in Multilingual Language Models: A~Comparative Analysis of mT5 and ByT5}
\author{Anh Dang \\
  MPI for Psycholinguistics\\
  Radboud University\\
  Utrecht University\\
  \texttt{t.t.a.dang@uu.nl} \\\And
  Limor Raviv \\
  MPI for Psycholinguistics \\
  University of Glasgow \\
  \texttt{limor.raviv@mpi.nl} \\\And
  Lukas Galke \\
  University of Southern Denmark\\
  MPI for Psycholinguistics\\
  \texttt{galke@imada.sdu.dk}}
\begin{document}
\maketitle
\begin{abstract}
Morphology is a crucial factor for multilingual language modeling as it poses direct challenges for tokenization. Here, we seek to understand how tokenization influences the morphological knowledge encoded in multilingual language models. Specifically, we capture the impact of tokenization by contrasting two multilingual language models: mT5 and ByT5. The two models share the same architecture, training objective, and training data and only differ in their tokenization strategies: subword tokenization vs.\@ character-level tokenization. Probing the morphological knowledge encoded in these models on four tasks and 17 languages, our analyses show that the models learn the morphological systems of some languages better than others and that morphological information is encoded in the middle and late layers. Finally, we show that languages with more irregularities benefit more from having a higher share of the pre-training data.
\end{abstract}

\section{Introduction}\label{sec:intro}

\begin{figure*}
    \centering
    \includegraphics[width=0.95\textwidth]{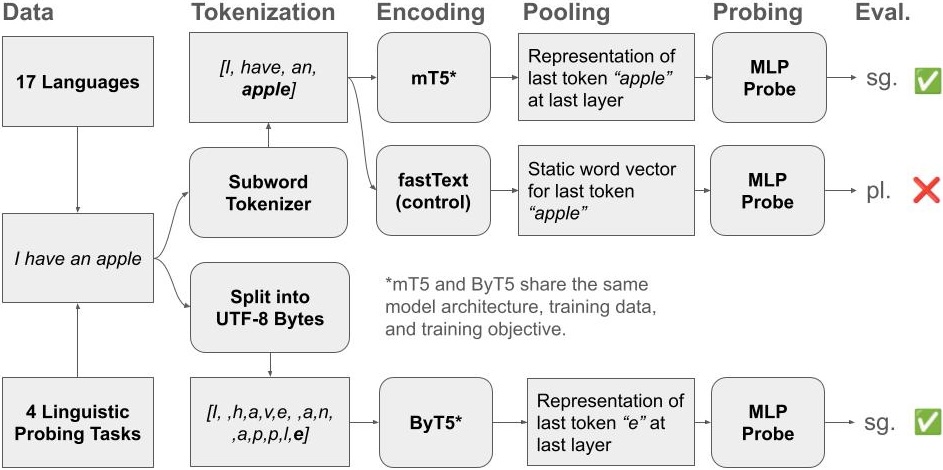}
    \caption{Overview of our experimental procedure}
    \label{fig:one}
\end{figure*}

Tokenization, the process of segmenting a text into individual units, plays a special role in language modeling as it is disconnected from the otherwise end-to-end training procedure~\cite{xue-etal-2021-mt5,xue2022byt5,sennrich2016neural}. Languages differ in their morphological structure~\cite{goldman2022morphology,dryer2013wals,evans2009myth,ackerman2013morphological} and it has been shown that morphologically more complex languages are harder to acquire by humans \citep{dekeyser2005makes, raviv2021makes, kempe2008second} and deep neural network models \citep{galke2023makes,park2021morphology, mielke2019kind,cotterell2018all}. Here, we seek to understand to what extent different tokenization strategies influence the ability of multilingual language models to capture morphological knowledge in different languages (See Figure~\ref{fig:one}).

Ideally, a language model would be equally proficient in a variety of languages~\cite{lample2019cross,conneau2020unsupervised,ruder2019survey}.
Understanding the influence of tokenization is crucial in the context of multilingual language modeling~\cite{xue-etal-2021-mt5,xue2022byt5,warstadt2020learning}, as it is challenging to find a set of tokens that is equally good for modeling all the languages in the world.
Beyond the proportions of languages in the training data, it is important to take into account the morphological structure of the different languages~\cite{anh-etal-2024-morphology,galke2023makes, cotterell2018all}. Importantly this needs to be already considered when selecting the data for learning the tokenizer -- even before language model pre-training -- as this influences what subword structures end up as the tokens to be processed by the language model. The issue of tokenization and, in related matter, how to mix different languages, are particularly relevant in the current era of large language models~\cite{touvron2023llama,brown2020language,bubeck2023sparks,wei2022emergent}, when aiming for similar performance on a diverse range of languages~\cite{le2023bloom}.

With models such as ByT5~\cite{xue2022byt5}, a character-level language model based on the T5 architecture~\cite{raffel2020exploring}, it has been shown that tokenizer-free language models yield commensurate downstream performance with their tokenizer-based counterparts~\cite{xue2022byt5,edman2024character}, such as  mT5~\cite{xue-etal-2021-mt5}, another T5-based model that trained on the exact same data as ByT5. Specifically, in machine translation, \citet{edman2024character} have found that character-level ByT5 yields similar performance as mT5 when allowing more training to recover word-level structures. Yet, the interplay of morphology and tokenization is so far poorly understood.

Here, we seek to dissect the root of these findings through analyzing the effect of the tokenization strategy (character-level vs.\@ subword-level) on the morphological knowledge encoded in contextualized representations of multilingual language models.
Specifically, we contrast ByT5 as a character-level multilingual language model and mT5 as a subword-level multilingual language model, with both models sharing the same architecture and being trained on the same data.
We use well-established structural probing techniques to capture the amount of linguistic information encoded in the contextualized word representations of the language models~\cite{rogers2021primer,manning2020emergent,belinkov2020linguistic,tenney2019bert}.
Focusing on morphology, we probe the contextualized representations of ByT5 and mT5 for morphological knowledge on 17 languages from a large-scale multilingual dataset~\cite{acs2023morphosyntactic}. In addition, we use the non-contextualized fastText model~\cite{bojanowski2017enriching} as a control to understand to what extent the context is important for multilingual language models reflecting morphosyntactic structures in their representations.
We further explore to what extent the captured morphological knowledge depends on the share that the languages have in the pre-training data, various linguistic factors, such as the type of task (number, tense, case, gender), and the language's degree of  morphological complexity, as quantified by the degree of irregularity~\cite{wu2019morphological} and type-to-token ratio (TTR) ~\cite{bentz2015adaptive}.

By systematically contrasting tokenizer-free ByT5 and subword-tokenized mT5, two pre-trained multilingual language models based on the T5 architecture, and trained on the same data, we find:

\begin{itemize}
\item Multilingual language models learn the morphological systems of some languages better than others.

\item Morphological information is encoded in the middle and late layers of the model.

\item Morphology is learned in earlier layers by models employing a standard tokenizer, yet character-level models display commensurate morphological knowledge at later layers.

\item The degree of irregularity plays a substantial role for capturing morphological knowledge, suggesting that more irregular languages would benefit from a higher proportion of training data.
\end{itemize}

\section{Background and Related Work}\label{sec:rw}

\paragraph{Morphology}
Morphology concerns how meaningful word units can be combined to express a range of grammatical information~\cite{bloomfield1933language}. Languages differ greatly in their morphological systems and degrees of morphological complexity \citep{dryer2013wals,lupyanLanguageStructurePartly2010,bloomfield1933language}. It has been shown that morphologically more complex languages are harder to acquire by humans \citep{dekeyser2005makes, raviv2021makes, kempe2008second} and deep neural networks~\citep{galke2023makes,cotterell2018all}.
Some studies specifically investigate the link between morphological complexity and the challenges in language modeling~\citep{cotterell2018all, mielke2019kind, park2021morphology, galke2023makes,anh-etal-2024-morphology}. However, the findings have been mixed. As most of these studies focus more on the overall learnability~\cite{park2021morphology,gerz2018language}, there may be confounds other than morphological complexity, which we isolate here with a fine-grained analysis.

\paragraph{Linguistic Probing}
Linguistic probing can be categorized into behavioral probing and structural probing~\cite{madsen2022post}.
Behavioral probing aims to understand how a language model behaves in a specific or new setting~\cite{hupkesTaxonomyReviewGeneralization2023b}.
Structural probing instead seeks to localize where linguistic abilities are encoded~\cite{rogers2021primer}.
It was suggested that morphological knowledge is mainly encoded at the lower layers~\citep{peters2018dissecting, belinkov2020linguistic}. However, most studies focus on analyzing monolingual language models, such as BERT~\cite{devlin2018bert}, or neural machine translation systems trained on pairs of languages~\citep{acs2023morphosyntactic,belinkov2020linguistic, bisazza2018lazy,edmiston2020systematic}, while studies on multilingual language models are rare.

\paragraph{Tokenization}
Tokenization is a crucial step in language modeling, especially for multilingual language models. The dominant tokenization methods for language models are based on the Byte-Pair Encoding (BPE) algorithm~\cite{sennrich2016neural}, and rely on the data mix to ensure coverage of different languages~\cite{le2023bloom}. Starting with single characters, BPE iteratively merges tokens based on co-occurrence statistics, allowing for both subword and multi-word tokens.
Comparing BPE with other subword tokenizers, \citet{ali2023tokenizer} found that the preference for tokenization methods differs across languages. While BPE works better for Germanic languages, such as German and English, Unigram~\citep{kudo-2018-subword} is more well-suited for Romance languages, such as Spanish.
The importance of
tokenization in multilingual language modeling is evident~\citep{hofmann2021superbizarre, toporkov2024role}, as many studies
propose new tokenization algorithms to make language models capture the
morphological structure of the different languages~\citep{goldman2022morphology}.

\paragraph{Character-level Language Models}
A line of research investigates character-level language models to circumvent the issues of multi-lingual tokenization~\citep{fleshman2023toucan,clark2022canine, xue2022byt5, gao-etal-2020-character, chung2016character, lee2017fully, kim2016character}.
In machine translation, \citet{lee2017fully} found that character-level tokenizers perform as well as or better than models based on sub-word tokenization. They highlighted that character-level tokenization offers better translation quality in multilingual and low-resource settings because of the shared vocabulary.
\citet{edman2024character} argued that character-level models are better at learning information that operates at a low level of granularity, such as morphology. Comparing translation capability between multilingual language models with standard tokenization (mT5) and character-level model (ByT5), they found that ByT5 outperforms mT5 in several aspects. First, ByT5 produces higher-quality translations than mT5, even in the case of low-resource languages. ByT5 is also better in handling rare and similar words.
In addition, low-resource
languages may benefit from shared vocabulary, namely the set of characters
\citep{gao-etal-2020-character}. However, it has also been suggested that the
effect varies across languages \citep{ali2023tokenizer}, which in-turn motivates the present study.

\paragraph{Summary}
Overall, earlier work on probing the morphological knowledge of language models is mainly conducted on machine translation and recurrent neural networks~\citep{belinkov2020linguistic, vylomova2017word}. More recently, the focus has been extended to Transformer-based language models~\citep{acs2023morphosyntactic, edmiston2020systematic, wu2020all}, yet the majority of studies investigated BERT and its variants~\cite{rogers2021primer}. Previous studies provide mixed findings about the morphological knowledge of language models and only very few multilingual studies.
Here, we complement the literature by analyzing how captured morphological knowledge is influenced by the tokenization strategy, the languages' proportions in the model's training data, the language's morphological complexity, and the task type.

\section{Models and Data} \label{sec:methods}

We compare two pre-trained language models: mT5 and ByT5. The two models share the same architecture and are trained on the same data with the same training objective (masked span prediction). The key difference between mT5 and ByT5 is their tokenization strategy: mT5 uses a standard subword tokenization strategy, whereas ByT5 operatores on character level. Thus, we can investigate the effect of tokenization.

\paragraph{mT5}
The mT5 model~\cite{xue-etal-2021-mt5} is a multilingual version of T5, an encoder-decoder language model \citep{raffel2020exploring}. It is trained on the mC4 corpus \citep{xue-etal-2021-mt5}, which consists of text in 101 languages compiled from the Common Crawl web scrape. Like T5, mT5 employs a masked span prediction training objective and the SentencePiece tokenizer \citep{kudo2018sentencepiece}, a variant of the BPE tokenizer by \citet{sennrich2016neural}. The vocabulary comprises approximately 250,000 subwords, covering 104 languages \citep{xue-etal-2021-mt5} while sharing (sub-)words between languages.

\paragraph{ByT5}
ByT5~\cite{xue2022byt5} is a tokenizer-free variant of mT5, inheriting most of the properties of mT5, using the same T5 model architecture, and the same masked span prediction training objective.  The only crucial difference between them is the tokenization method: While mT5 uses a standard tokenizer, ByT5 operates on single-character tokens, or more precise: UTF-8 encoded bytes.  Another difference is that ByT5's encoder stack consists of three times more layers than the decoder to process a larger number of bytes.
For the masked span prediction objective, ByT5 uses a span of 20 bytes.
The similarity in architecture and sizes of mT5 and ByT5 enables our comparison of how the tokenization strategy impacts the morphological knowledge encoded in the learned representations.

\paragraph{Dataset}
We use the multilingual morphological probing dataset by \citet{acs2023morphosyntactic}. The dataset consists of 247 probing tasks, available in 42 languages and 10 language families. It is built upon the Universal Dependencies tree bank and covers both frequent and infrequent words. In each language, each task includes a training set of 2000 examples, a development test of 200 examples, and a test set of 200 examples. We selected 16 out of all 42 languages which had at least two tasks available. However, we also included Arabic despite having only one task available to better cover the Semitic language family. In total, our considered dataset consists of 17 languages and 43 morphological probing tasks, covering number, case, gender, and tense -- focusing on nominal and verbal inflection.

\section{Methodology}
\paragraph{Feature Extraction}\label{extract}
For training, we extracted the contextualized embeddings of the words in the training set for each task in each language and each model. Both mT5 and ByT5 are available in different sizes. We chose to test the \texttt{mT5-base} model. We froze the weights and extracted the hidden states for the entire sentence before extracting the word embedding corresponding to the target. Since we also aim to look at how much morphological knowledge is learned at each layer, we extracted the word embedding at each hidden layer of the network, including the input embedding layer. Each word representation is associated with a label, which corresponds to the respective morphological feature from the task.

\paragraph{Probing Classifiers}\label{probe}
Previous studies often use two architectures for probing classifiers, namely linear classifiers \citep{hupkes2018visualisation, belinkov2020linguistic} and multilayer perceptrons (MLPs) \citep{lin2019open, conneau2018you, adi2016fine, ettinger2018assessing, zhang2018language}. Both types of probes have received convincing arguments. Linear probes capture the information that can be straightforwardly detected in the representations, thus providing a faithful indication of their linguistic knowledge~\citep{liu2019linguistic, belinkov2020linguistic}, whereas MLP probes can capture all information encoded in the representations, including information that requires nonlinear processing~\citep{hewitt2019designing}. studies show that linear classifiers and MLPs produce similar accuracy.\citep{conneau2018you, belinkov2017neural, qian2016investigating}. We have considered both types of probes for this study (see Appendix~\ref{app:probing}). For the main results, we employ MLP probes.

\paragraph{Subword Pooling}
In both mT5 and ByT5, words are segmented into either subword units or characters. As such, when passing through the hidden layers, each subword or character has its own embedding. There are several methods to then approximate the embedding for an entire word: The first method is to take the weighted \textbf{average} of the embeddings of all components. The second way is to consider the embedding of the \textbf{last} subword or character as the representation for the entire word. Both methods have limitations. Averaging the token embeddings may cancel out some information, while the last embedding may not contain all the information about the entire word. \citet{belinkov2020linguistic} compared both methods and found that using the embeddings of the last token produced higher accuracy scores. We have considered both options (see Appendix~\ref{app:pooling}) and can confirm that last-token pooling leads to higher probing accuracy.
For the main results, we employ last-token pooling.

\paragraph{Evaluation and Control} \label{control}
We aim to probe the morphological knowledge of a range of typologically different languages. While the dataset of \citet{acs2023morphosyntactic} supports 42 languages, in some languages, there is a large gap between the number of tasks in each language. While Russian has 12 tasks, Polish and Armenian have only one task. To ensure a fair comparison of the learned morphological representations, we selected languages with at least two tasks along with Arabic to cover the Semitic language family.
We focused exclusively on the morphological properties of words, excluding agreement tasks.
In total, we ran 43 morphological probing tasks for 17 languages.
Appendix~\ref{app:languages}
provides the details of the morphological properties of the 17 considered languages, their morphological complexity scores, and their proportion in the ByT5/mT5 training data.
Appendix~\ref{app:tasks} provides a detailed description of the types of probing tasks, covering number, case, gender, and tense.

As non-contextualized control, we employ fastText word embeddings~\citep{bojanowski2017enriching} which are available in 157 languages. We probed fastText embeddings using the exact same procedure and evaluation metric. We obtained the static word embeddings without any pooling. Contrasting the contextualized representations of mT5 and ByT5 with fastText allows us to quantify to what extend the contextualized models make use the sentence context to tackle the morphological tasks.

\section{Results}\label{sec:results}
\paragraph{Overall Probing Accuracy}
We first looked at the overall probing performance of mT5, ByT5, and fastText as well as the differences between the two Transformer-based models compared to the fastText baseline. For all analyses except for the layer-wise analysis, we used the probing accuracy of the last hidden layer. To obtain the overall performance of mT5 and ByT5, we averaged over all languages and tasks, resulting in a single accuracy score for each model (see Table~\ref{tab:acc-mean}). Full results for each task and language are provided in Appendix~\ref{app:extended-results}.

\begin{table}[ht]
    \small
    \centering
    \begin{tabular}{lc}
    \toprule
      \textbf{Model} & \textbf{Mean Probing Accuracy}  \\ [0.5ex]
      \midrule
      mT5-base & 82.57 \\
      ByT5-base & \textbf{82.86} \\
      fastText (baseline) & 77.52 \\ [0.5ex]
      \bottomrule
    \end{tabular}
    \caption{Probing accuracy of mT5, ByT5 and fastText, averaged over languages and tasks}\label{tab:acc-mean}
\end{table}

On the surface, it appears that mT5 and ByT5 have comparable performance and both models outperformed fastText. ByT5 slightly surpassed mT5, yet this difference is very small. This finding is different from that of \citet{belinkov2017neural}, who found that character-level tokenizers are better than subword tokenizers in representing morphology. Considering the differences between both LLMs and the fastText baseline, as shown in Table~\ref{tab:acc-mean-lang}, it can be seen that they generally outperformed the baseline yet perform substantially worse than baseline in French and Russian. In addition, mT5 achieved lower probing accuracy than fastText in Arabic and Hindi. ByT5 performed worse than fastText in German and Urdu.

\begin{table}[hbt!]
    \small
    \centering
    \begin{tabular}{lcccc}
    \toprule
      \textbf{Language} & \textbf{Family} & \multicolumn{3}{c}{\textbf{Model}}\\
      \midrule
       & &  \textbf{mT5} & \textbf{ByT5} & \textbf{fastText} \\

      \midrule
       English & Germanic & 98.00 & \textbf{98.50}  & 97.75  \\
       Dutch & Germanic &  \textbf{93.50}   &  92.75   &  82.25   \\
       German & Germanic   & 68.46  &  \textbf{72.40} & 68.87  \\
       \midrule
       French & Romance  & 71.93    & 77.53   &  \textbf{92.95}  \\
       Spanish & Romance   &  93.83   & \textbf{94.00}  & 71.16    \\
       Portuguese & Romance  &  95.00   &  \textbf{96.25}  & 88.16  \\
       Romanian & Romance  &  94.75    &  \textbf{95.25}   & 92.25  \\
       \midrule
       Hebrew & Semitic   & 97.50  & \textbf{98.00}  & 92.49  \\
       Arabic & Semitic   &  \textbf{66.17}    &  49.25  & 37.81  \\
       \midrule
       Russian & Slavic  & 61.26   & 57.43  & \textbf{79.20}  \\
       Czech & Slavic  & \textbf{88.79}   &  87.79  &  78.81  \\
       \midrule
       Hindi & Indic   & 67.83 & \textbf{87.50}  & 58.02  \\
       Urdu & Indic    &  \textbf{88.50}   &  84.50  &
       74.33  \\
       \midrule
       Turkish & Turkic   & \textbf{94.78}   &  83.69  & 78.28  \\
       \midrule
       Latvian & Baltic   & \textbf{77.14}  &  72.01  & 73.76  \\
       \midrule
       Estonian & Uralic    &  \textbf{91.08}  &  90.03  &  82.94  \\
       \midrule
       Basque & Basque   & 91.69 & \textbf{91.91}   &  52.60  \\ [0.5ex]
      \bottomrule
    \end{tabular}
    \caption{Probing accuracy of mT5, ByT5, and fastText by languages, language families, averaged over tasks. The best score per language is marked in bold font.}
    \label{tab:acc-mean-lang}
\end{table}

Table~\ref{tab:acc-mean-lang} shows the difference between accuracy scores of mT5 and ByT5, grouped by language. It can be seen that accuracy scores are not equal across both languages and tasks. Comparing mT5 and ByT5, it seems that they perform on par with each other in most languages. However, mT5 scores higher in Turkish and ByT5 achieve much higher on Hindi tasks.
The difference between contextualized language models and non-contextualized fastText also tells to what extent contextual information affects morphological abilities.
From the results, we observed that for most languages, mT5 and ByT5 achieved considerably higher probing accuracy than the baseline. The largest difference is observed in the case of Basque (> 50\%). This implies that contextual word embeddings capture morphological knowledge better than static embeddings for these languages. In contrast, the results are lower than the non-contextual baseline in French and Russian. Contextual information seems to make accessing morphological information more difficult in these two languages.

\begin{table}[ht]
   \small
    \centering
    \begin{tabular}{lccc}
    \toprule
      \textbf{Task} & \textbf{mT5} & \textbf{ByT5} & \textbf{fastText} \\ [0.5ex]
      \midrule
      Number   & \textbf{93.75}  & 93.56 & 88.15 \\
      Tense   &  80.30 & 72.44 & \textbf{80.89} \\
      Gender   &  76.85  & \textbf{85.16} & 77.61 \\
      Case & \textbf{71.53} & 69.16 & 55.49 \\ [0.5ex]
      \bottomrule
      \end{tabular}
    \caption{Probing accuracy of mT5, ByT5, and fastText, averaged over languages and tasks}\label{tab:acc-task}
\end{table}

 Table~\ref{tab:acc-task} shows the probing accuracy scores for each language, averaged over tasks. It appears that there are some differences in accuracy across languages, with the hardest language being Russian for mT5 and Arabic for ByT5. The accuracy scores of some languages are higher than the others. Unsurprisingly, the models perform best on the English language, followed by Hebrew, Portuguese, and Romanian. The models learn the morphological systems of German, French, Estonian, and Latvian moderately well. Arabic and Russian achieved lowest accuracy scores. However, Arabic results should be interpreted with caution as there is only one task available (i.e., case).

\paragraph{Layer-wise Analysis}

\begin{figure*}[hbt!]
    \centering
    \includegraphics[width=0.45\textwidth]{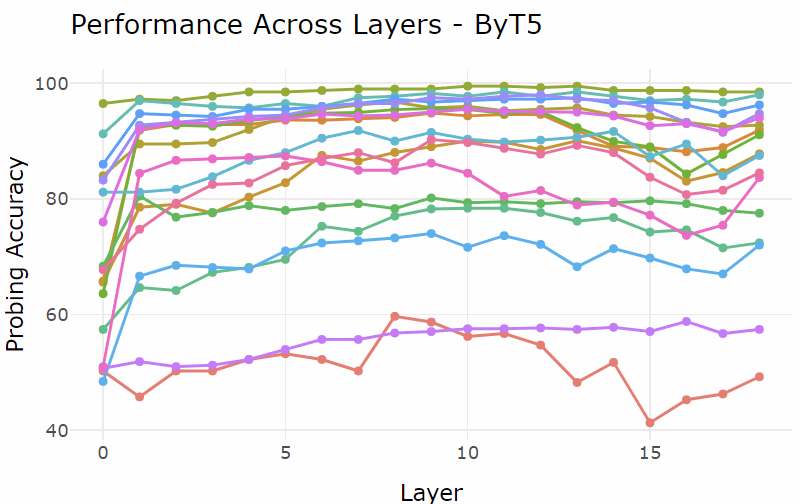}
    \includegraphics[width=0.45\textwidth]{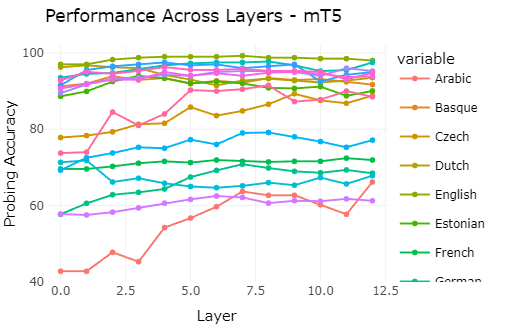}
    \includegraphics[width=0.45\textwidth]{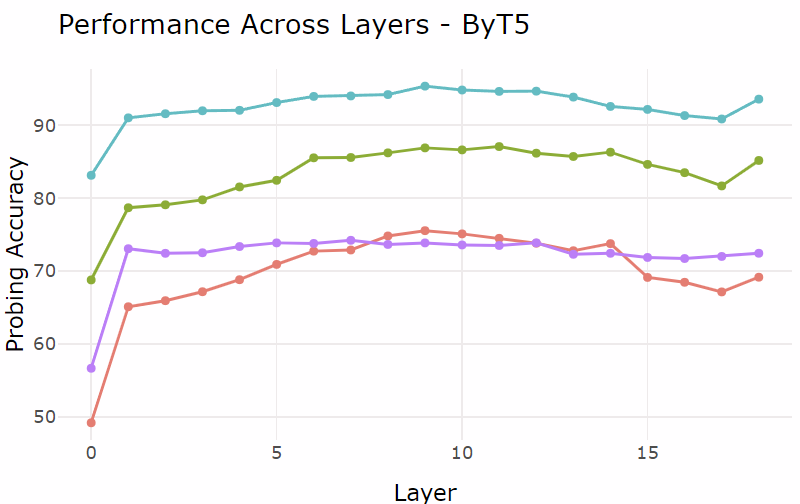}
    \includegraphics[width=0.45\textwidth]{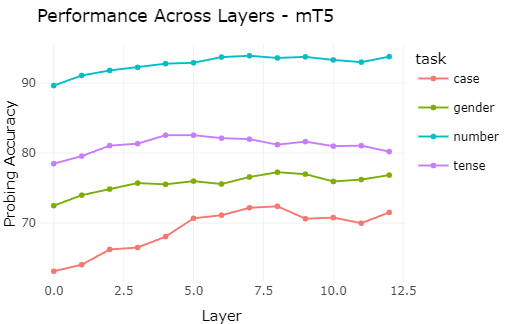}

    \caption{Probing accuracy of ByT5 (left) and mT5 (right) across layers grouped by languages and tasks. Each line represents a language (top) or a task (bottom). Each data point is the accuracy scores at each layer of each language.}\label{fig:plot}
\end{figure*}

Figure~\ref{fig:plot} illustrates the difference in probing accuracy between languages and between tasks for mT5 and ByT5. It can be seen that there are some degrees of variation between layers. In languages that show high overall performance, namely English, Dutch, Portuguese, Spanish, Basque, and Hebrew, probing accuracy shows very little improvement over layers. In other languages, accuracy increases, reaches its peak at the middle and slightly decreases at late layers in other languages. This finding is partly consistent with \citet{acs2023morphosyntactic}, \citet{edmiston2020systematic}, and \citet{hewitt2021conditional}, who also reported best performance in the middle to late layers. However, we further show that this is not true for all languages. There are cases where morphological knowledge is successfully learned in the early layer and carried on throughout the network. Our results contrast with \citet{belinkov2020linguistic} and \citet{peters2018dissecting}, who found that morphological information is best encoded in the first layer of the model, and then has the tendency to decrease over time.

Comparing mT5 and ByT5, there are a few noticeable differences. In the plots for ByT5, accuracy scores of each language and each task improves considerably after the embedding layer. This trend is less visible in mT5, although performance does improve over layers. Morphology is better learned in the embedding layer of mT5 than that of ByT5.  This may imply that character-level language models need more layers to capture morphological patterns of languages.

\paragraph{Effect of Task Type}

To investigate whether morphological features are learned differently by mT5 and ByT5, we averaged the scores for each task across all languages, resulting in a single score for each task (see Table~\ref{tab:acc-task}). The results strongly suggest that each morphological feature is encoded differently. Interestingly, mT5 and ByT5 show different patterns. Both models perform equally well at number and worst at case. However, tense is learned better than gender by mT5 while the opposite is true for ByT5. Comparing both language models with the baseline, they surpass the baseline in all tasks except for tense, where ByT5 performs considerably worse than fastText.

Case seems to be the hardest task for both models. Besides the case task often having more possible classes than other features, case is also more context-dependent than other features. Case marking is used to indicate the syntactic function of the word in the sentence. As such, one word may have different cases in different contexts and thus is inflected distinctively. Gender is also relatively difficult, especially for mT5. However, looking at individual languages (see Table~\ref{tab:acc-mean-lang}), the mean score of mT5 is affected by Hindi and Latvian, whose scores are exceptionally lower than the baseline (less than 25\%). Except for those two languages, mT5 and ByT5 perform equally well.

\paragraph{Morphological Complexity and Training Data}

We explore the effects of two types of morphological complexity, namely TTR~\cite{bentz2015adaptive} and the degree of irregularity~\cite{wu2019morphological} on probing accuracy. Complexity values per language can be found in Appendix~\ref{app:languages}.
We hypothesize that probing accuracy is influenced by the proportion of the respective language in the training data, which then also modulates the effect of a language's morphological complexity.
To understand the effect of these factors and their interaction, we fitted a generalized mixed effect logistic regression model predicting accuracy from the two morphological complexity measures and the proportion of training data. There are 200 data points for each task in each language. The analysis was conducted with the R-package \texttt{lme4}. All variables were scaled and centered. Random intercepts were added for language and task.

Our results show significant positive effects of training data size and irregularity on probing accuracy, as well as their interaction.
In more detail, there is a strong effect of the amount of training data and a language's degree of irregularity on probing accuracy (training data: $\beta$ = 2.11, \textit{SE} = .55, \textit{p} < .001, irregularity: $\beta$ = 2.83, \textit{SE} = .42, \textit{p} < .001).
The effects of training data and irregularity were highly correlated (0.831).
In addition, there is an interaction effect between the language's irregularity and its training data in the model ($\beta$ = 2.03, \textit{SE} = .31, \textit{p} < .001). The effect of training data size on probing accuracy is stronger when there is more irregularity in the language. This means that high irregularity in the morphological systems amplifies the impact of training data on the morphological abilities of language models.
Detailed results of the statistical models can be found in Appendix~\ref{app:stats}.

\section{Discussion}\label{sec:discussion}

\textbf{Languages' morphology is learned differently} Our probing results in mT5 and ByT5 show that the morphological knowledge of some languages is better represented than the others. Some languages (e.g., English, Dutch) achieved nearly perfect accuracy in probing tasks (higher than 90\%). However, both mT5 and ByT5 performed worse at German, French, Russian, and Arabic tasks. These results to some extent contradict \citet{edmiston2020systematic}, who found comparable performance in all languages. \citet{acs2023morphosyntactic} also did not observe performance differences across languages, but only between part-of-speech and morphological features.

\textbf{Differences across tasks} We observed that some tasks are more difficult for the language models: Number is the easiest task whereas case is the hardest one. This difference can be partly attributed to the higher number of possible categories but also to the context-dependent nature of case. This is supported by the non-contextulaized baseline results for case, which are substantially lower than both mT5 and ByT5. These findings are in agreement with earler findings by \citet{bisazza2018lazy} and \citet{edmiston2020systematic}. Why tense and also case features are particularly challenging to find in the representations of character-based language models is an interesting question for future research.

\textbf{Character-level models are on par with sub-word level models in representing morphology}
We found that both models yield highly similar performance on the probing tasks. Remarkably, despite being tokenized at the byte level, ByT5 captures morphological knowledge in the early layers of the network, on par with mT5. We found that morphology is learned later in ByT5 than in mT5. The gap between the embedding layer and the first layer is considerably higher in ByT5. Although mT5 and ByT5 eventually capture a similar amount of morphological knowledge, ByT5 needs more layers for processing the input to yield results that are comparable with mT5. \citet{belinkov2020linguistic} and \citet{vylomova2017word} tested machine translation systems and found character-level tokenizers to surpass BPE in learning morphology. However, through investigating large-scale language models, we found no advantage of character-level tokenizer in encoding morphology. We found that subword tokenization works better for Turkish, whereas character-level tokenization benefits Hindi morphology -- highlighting that the preferable model differs across languages.

\textbf{Morphology is best represented in middle to late layers} Our findings show that morphological knowledge generally improves over layers in both language models. There are languages which have high performance across layers. Yet, ByT5 shows greater improvement after the embedding layer than mT5. We also observed that morphological information is best encoded in the middle to late layers of the models in some languages. This finding supports \citet{acs2023morphosyntactic}, \citet{hewitt2021conditional}, and \citet{edmiston2020systematic}. Our findings differ from \citet{belinkov2017evaluating, tenney2019bert, peters2018dissecting}, who found that morphology is a low-level feature and is encoded along with word identity in the first layer of the network.

\textbf{Morphological irregularity amplifies the effect of training data}
Considering the relationship between morphological knowledge encoded in language models and the languages' morphological complexity, our analysis reveals effects of morphological irregularity and training data sizes on the performance of probing classifiers, in a way that the effect of irregularity is mediated by training data. When a language is highly irregular, a larger share of the training data is beneficial to fully capture its morphological system.
Previous studies on the effect of training data sizes show that its effect is not present at the representation level yet at the downstream level~\citep{warstadt2020learning, zhang2018language}.
We have shown here that the importance of the relative training data size in multilingual language modeling can be already found when probing for morphological knowledge.
Considering the interplay with morphological complexity, \citet{mielke2019kind} and \citet{gerz2018language} correlated modeling difficulty with morphological counting complexity~\citep{sagot2013comparing}, vocabulary sizes of languages, and dependency length -- and found vocabulary size to be the most important factor. Our study complements those findings by establishing that the degree of irregularity plays a substantial role for what morphological properties are captured by a language model -- and that this factor amplifies the effect of the language's share of the pre-training data.

It may seem unexpected that a higher degree of irregularity has a positive effect on probing accuracy. However, a possible explanation is that irregular forms are better memorized \emph{because} they appear more often in the training data, given the correlation of irregularity with frequency~\cite{wu2019morphological}. This could also be linked to the word predictability advantage of Zipfian distributions that has been shown to aid word segmentation in humans~\cite{lavi2022learnability} -- which we deem an interesting direction for future work.

\section{Conclusion}
We have analyzed the effects of tokenization, training data proportions, and linguistic factors on morphological knowledge encoded in the parameters of pre-trained multilingual language models. Through analyzing 17 languages and 4 morphological tasks, we have shown that the morphological knowledge encoded in multilingual language models differs across languages, despite the global average scores being similar. Beyond differences across languages, we also found differences across tasks, showing that tense and case are particularly hard to find in the representations of character-based language models. To further understand what exactly influences those difference, we have analyzed the effect of morphological complexity in relation to the language's proportion in the language model's pre-training data. We found that the degree of irregularity plays a significant role and amplifies the effect of training data, suggesting that more irregular languages benefit from a having a higher share in the data mix used for pre-training.

\section{Limitations} Several limitations should be taken into account: First, our experiments do not take orthographic transparency into account, as all models and complexity measures are based on written language. Second, we ran each experiment only once due to the high volume of experiments. Third, while we aimed to cover as many typologically different languages as possible, the dataset we use supports mostly Indo-European languages, such that 12 out of our 17 considered languages are Indo-European. Moreover, the current study only focuses on inflectional morphology due to the lack of datasets for probing derivational morphology. Lastly, p-values should be treated with care when the sample size (here: 17,200) is large~\citep[][]{sogaard2014s}.

\section{Ethical Considerations}
We emphasize that morphological complexity of languages bears no implication on their quality -- having more complexity does not make one language better than another~\citep[see][]{raviv2022simple}.

\bibliography{references}

\appendix

\section{Effect of Probing Architecture}\label{app:probing}

Previous studies on probing linguistic features have had a debate over which type of probe is sufficient to extract the relevant knowledge, but insufficient to learn the knowledge itself \citep{belinkov2022probing}. Here, we also compare the accuracy scores of MLP probes and linear probes for mT5 and ByT5, as shown in Table ~\ref{tab:acc-linear}.

\begin{table}[htbp]
    \centering
    \begin{tabular}{lcc}
    \toprule
      \textbf{Model} & \multicolumn{2}{c}{\textbf{Probing Architecture}}  \\ [0.5ex]
      \midrule
          & \textbf{MLP} &  \textbf{Linear} \\

      \midrule
      mT5-base & 82.57 & 80.37 \\
      ByT5-base & 82.86 & 80.61 \\ [0.5ex]
      \bottomrule
    \end{tabular}
    \caption{Probing accuracy of mT5 and ByT5 when using MLP and linear classifiers, averaged over languages and tasks}\label{tab:acc-linear}
\end{table}

We observed no considerable differences in accuracy scores across types of probes. For both types of probes, the mean accuracy scores are all around  80\%. This pattern is most inline with \citet{acs2023morphosyntactic}, suggesting that linear classifiers are as effective as non-linear ones in extracting morphological knowledge of multilingual LLMs. Moreover, the observation that linear probes perform equally well as MLPs implies that morphology is a relatively simple feature that can be learned early and straightforwardly by the models.

\section{Effect of Pooling Methods}\label{app:pooling}

We compare the probing accuracy when using these two pooling methods, namely the \textbf{average} and \textbf{last} method. Table ~\ref{tab:acc-pooling} shows the results of when using these two methods for mT5 and ByT5.

\begin{table}[ht]
    \centering
    \begin{tabular}{lcc}
    \toprule
      \textbf{Model} & \multicolumn{2}{c}{\textbf{Subword Pooling Method}}  \\ [0.5ex]
      \midrule
          & \textbf{last} &  \textbf{average} \\

      \midrule
      mT5-base & 82.57 & 75.28 \\
      ByT5-base & 82.86 & 76.40 \\ [0.5ex]
      \bottomrule
    \end{tabular}
    \caption{Probing accuracy of mT5 and ByT5 when using different subword pooling methods. The results were averaged over languages and tasks and the reported probes are MLPs}\label{tab:acc-pooling}
\end{table}

Comparing the accuracy scores of the two pooling methods, it can be seen that the \textit{last} method achieved considerably higher accuracy scores than the \textit{average} method. The difference is approximately 6-7 points. This also holds for both models. It seems that the representational information and/or the morphological content of a word is mostly encoded in its last token. Previous comparisons have also reported similar results \citep{acs2023morphosyntactic, acs2021subword, belinkov2020linguistic}.

\section{Details of the Considered Languages}\label{app:languages}

\subsection{Morphological Properties}
Table~\ref{tab:morphprop} shows the morphological properties of the considered languages.
\begin{table*}[ht]
    \centering
    \caption{Morphological properties of 17 investigated languages (N = noun; V = verb)}\label{tab:morphprop}
    \begin{tabular}{lcccccc}
    \toprule
        \textbf{\small{Language}} &\textbf{\small{Family}} & \textbf{\small{POS}} & \textbf{\small{Number}} & \textbf{\small{Tense}} & \textbf{\small{Case}} & \textbf{\small{Gender}} \\ [0.5ex]
        \midrule
        English & Germanic & N & 2 & -- & -- & -- \\
        English & Germanic & V & -- & -- & 2  & -- \\
        \midrule
        German & Germanic & N & 2 & -- & 4 & 3 \\
        German & Germanic & V & -- & 2  & -- & --  \\
        \midrule
        Dutch & Germanic & N &  2 & --  & -- & 2 \\
        \midrule
        French & Romance & N & 2 & -- & --& 2 \\
        French & Romance & V & -- & 4 &  -- & -- \\
        \midrule
        Spanish & Romance & N & 2 & -- & -- & 2 \\
        Spanish & Romance & V & -- & 4 & -- & -- \\
        \midrule
        Portuguese & Romance & N & 2 & -- & -- & 2 \\
        \midrule
        Romanian & Romance & N & 2 & -- & -- & 2 \\
        \midrule
        Turkish & Turkic & N & 2 & -- & 7 &  --\\
        \midrule
        Russian & Slavic & N & 2 & -- & 6  & 3\\
        Russian & Slavic & V & -- & 3 & -- & -- \\
        \midrule
        Czech & Slavic &  N & 2 & --  & -- & 3 \\
        \midrule
        Hebrew & Semitic & N & 2 & -- & -- &  2  \\
        \midrule
        Hindi & Indic & N & 2 & -- & 2 & 2 \\
        \midrule
        Urdu & Indic & N & 2 & -- & 2 & -- \\
        Urdu & Indic & V & 2 & -- & -- & -- \\
        \midrule
        Basque & Basque & N & 2 & -- & 11 & -- \\
        \midrule
        Estonian & Uralic & N & 2 & -- & 18 & -- \\
        Estonian & Uralic & V & -- & -- & -- & -- \\
        \midrule
        Latvian & Baltic & N & 3 & -- & 5 & 3 \\
        Latvian & Baltic & V & -- & 3 & -- & -- \\
        \midrule
        Arabic & Semitic & N & -- & -- & -- & -- \\   [1ex]
        \bottomrule
    \end{tabular}
\end{table*}

\subsection{Proportion of mT5/ByT5's training data and morphological complexity of the language}

Table~\ref{tab:language-detail} shows the language's proportion of training data and their morphological complexity scores: TTR and Irregularity.

\begin{table}[hbt!]
    \centering
    \begin{tabular}{lrrr}
        \toprule
         \textbf{Language}& \textbf{Tr. Data} & \textbf{TTR} & \textbf{Irregularity}  \\
         \midrule
         English   &  5.67\% & -0.460 & -5.94 \\
         German & 3.05\%  & -0.010 & -6.28 \\
         Dutch &  1.98\% & -0.390 & -6.68  \\
         French & 2.89\% & -0.340 & -4.16  \\
         Romanian & 1.58\% & -0.420 & -3.40 \\
         Spanish &  3.09\% & 0.001 & -8.81 \\
         Portuguese &  2.360\% & 0.038 & -9.11 \\
         Turkish & 1.93\%  & 1.550 & -5.96 \\
         Czech & 1.72\%  & 0.430 &-5.63 \\
         Russian & 3.71\%  & 0.870 & -7.74 \\
         Hebrew &  1.06\% & 2.020 & -1.78  \\
         Arabic & 1.66\%  & 1.630 & -0.06 \\
         Hindi & 1.21\%  & -0.300 & -2.10 \\
         Estonian & 0.89\%  & 1.760 & -2.79 \\
         Latvian & 0.87\%  & 0.770 & -7.90 \\
         Urdu & 0.61\%  & -0.450 & 9.20 \\
         Basque & 0.57\%  & 1.310 & 19.86 \\
         \bottomrule
    \end{tabular}
    \caption{Percentages of training data of mT5 and Byt5 from \citet{xue-etal-2021-mt5}, irregularity scores from \citet{wu2019morphological}, and TTR scores from \citet{bentz2015adaptive} for each investigated language. For irregularity, higher scores mean being more morphologically irregular. In contrast, higher TTRs mean higher complexity.}\label{tab:language-detail}
\end{table}

\section{Considered Probing Tasks}\label{app:tasks}

Here, we provides an overview the morphological properties that we investigate in the study, namely number, tense, case, and gender, and how they may vary between languages.
\cite{acs2023morphosyntactic}

\paragraph{Number} In many languages, especially inflected languages, nouns are marked as either singular or plural~\citep{bloomfield1933language}. An exception is Latvian, which includes singular, plural, and partitive nouns. Plurality is usually expressed by adding certain endings to the nouns, and sometimes include changing their vowels. These endings are determined in different ways across languages. In English, they are dependent on whether the nouns end with a consonant or vowel. However, in some Indo-European languages such as Spanish, the plural form of nouns is affected by their gender. In this task, the LLMs have to predict whether the target word is a plural or singular noun.

\paragraph{Tense}
Most languages mark tenses~\citep{bloomfield1933language}. In some languages, tenses are indicated by inflecting verbs. In other languages, for example, Estonian, adjectives can also express tense. In certain languages, tense can interact with other morphological features, namely mood and aspect. Inflection patterns for tense are usually dependent on the ending, conjugation pattern of the verbs, and whether they are regular or irregular. In Spanish and French, it is also dependent on the subject pronouns. In Hindi, verbs indicating tense must agree with gender and number of the subject.

\paragraph{Case} A case system is a grammatical category used in many languages to mark the relationship between a noun or pronoun and other words in a sentence. Case marking is typically indicated through inflection. The number of cases varies across languages. Cases are often marked with inflection. In some languages, case often affects how articles, pronouns, and adjectives should be inflected. Previous probing studies show that case is often one of the most challenging morphological categories to be learned by language models \citep{edmiston2020systematic, bisazza2018lazy, acs2023morphosyntactic}.

\paragraph{Gender} Most Indo-European languages mark genders in nouns and often require agreement with in other part-of-speech in the sentence, such as verbs and adjectives \citet{bloomfield1933language}. Gender systems exhibit substantial diversity in their number of genders, assignment rules. Some languages (e.g., Basque) do not have a gender system. Romance languages have a binary gender system. On the other hand, Germanic and Slavic languages often have more than two genders. For example, Dutch nouns are either common or neuter while German nouns can be masculine, feminine, or neuter. The gender carries little semantic information and is often characterized by its ending. Spanish masculine nouns end in \textit{"-o"} while feminine nouns end in \textit{"a"}. There is also a certain degree of irregularity.

\section{Extended Results}\label{app:extended-results}
Table~\ref{tab:acc-full} shows the full breakdown of probing accuracy per task and per language.
\begin{table*}
\centering
\caption{Accuracy scores of each task in each language from mT5, ByT5, and fastText}\label{tab:acc-full}
\begin{tabular}{cccccc}
  \toprule
   No. & \textbf{Language} & \textbf{Task} & \textbf{mT5} & \textbf{ByT5} & \textbf{fastText}   \\
  \midrule
  1 & Arabic & case & 66.17  & 49.25 &  37.81 \\
  \midrule
  2 & Basque & case & 91.39 & 92.55 & 17.22  \\
  3 & Basque & number & 92.00 & 89.42  & 87.99  \\
  \midrule
  4 & Czech &kgender & 81.09 & 78.87  &  72.13  \\
  5 & Czech & number & 96.50 & 90.05  & 85.50  \\
  \midrule
  6 & Dutch & gender & 90.00 & 88.39  &  72.50 \\
  7 & Dutch & number & 97.00 & 98.24  &  92.00 \\
  \midrule
  8 & English & number & 97.50 & 98.55  &  98.50 \\
  9 & English & tense & 98.50 & 98.53  &  97.00 \\
  \midrule
  10 & Estonian & case & 88.10 & 86.84  &  62.85 \\
  11 & Estonian & number & 91.50 & 92.47  & 89.49  \\
  12 & Estonian & tense & 90.50 & 93.84  & 96.49 \\
  \midrule
  13 & French & gender & 95.00 & 90.39  &  92.00 \\
  14 & French & number & 99.50 & 98.24  &  95.49 \\
  15 & French & tense & 21.29 & 46.30  & 91.08 \\
  \midrule
  16 & German & case & 65.00 & 40.42   &  28.00 \\
  17 & German & gender & 29.85 & 74.42  & 76.00  \\
  18 & German & number & 92.00  & 89.00   &  84.50 \\
  19 & German & tense & 87.00 & 85.95  &   87.00 \\
  \midrule
  20 & Hebrew & gender & 95.50 & 95.05   &  89.49 \\
  21 & Hebrew & number & 99.50 & 98.82  & 95.49  \\
  \midrule
  22 & Hindi & case & 13.00 & 81.58  & 63.49  \\
  23 & Hindi & gender & 95.50 & 90.74   & 49.00 \\
  24 & Hindi & number & 95.00 & 90.95  &  61.57 \\
  \midrule
  25 & Latvian & case & 89.50 & 98.74   & 84.50  \\
  26 & Latvian & gender & 68.00 & 32.61  &  64.49 \\
  \midrule
  27 & Latvian & number & 68.50 & 70.39  &  63.49 \\
  28 & Latvian & tense & 82.59 & 84.50   &  82.58 \\
  \midrule
  29 & Portuguese & gender & 95.00 & 93.66   &  97.00 \\
  30 & Portuguese & number & 95.00 & 97.55  & 97.50 \\
  \midrule
  31 & Romanian & gender & 93.50  & 93.68   & 91.00   \\
  32 & Romanian & number & 97.00 & 95.84  & 93.50 \\
  \midrule
  33 & Russian & case & 48.04  & 36.02   &  82.55  \\
  34 & Russian & gender & 8.46 & 82.98  &   50.74 \\
  35 & Russian & number & 96.00 & 93.39   &  91.50 \\
  36 & Russian & tense & 92.54 & 9.59  & 92.03  \\
  \midrule
  37 & Spanish & gender & 93.50 & 94.66   & 97.00  \\
  38 & Spanish & number & 99.00 & 97.84  &   97.50 \\
  39 & Spanish & tense & 89.00 & 86.47   & 31.00  \\
  \midrule
  40 & Turkish & case & 95.57 & 72.88   & 61.57   \\
  41 & Turkish & number & 94.00 & 89.32   & 94.99  \\
  \midrule
  42 & Urdu & case & 87.00 & 77.37  & 61.50 \\
  43 & Urdu & number & 90.00 & 90.87   & 81.49 \\
  \bottomrule
\end{tabular}
\end{table*}

\section{Detailed Results of the Statistical Analysis}\label{app:stats}
We provide the formulation and results of the statistical models. We define our model as the combination of two interaction terms between \texttt{irregularity} and \texttt{TD} and between \texttt{TTR} and \texttt{TD}. The syntax is as follows. The results can be found in Table~\ref{tab:model-m}.

\texttt{accuracy $\sim$ irregularity*TD + TTR*TD + (1|language) + (1|task)}

\begin{table*}[hbt!]
    \centering
    \caption{\small{Results of linear mixed effect regression with \textit{probing accuracy} of mT5 as outcome variable and \textit{language} and \textit{task} as random effects. Fixed effects are irregularity (I) and TTR (T), training data (TD) and their two-way interactions.}}\label{tab:model-m}
    \begin{tabular}{lllrl}
    \toprule
    \textbf{Fixed Effects} \\
    \midrule
    Variable &    Estimate &       SE  & \textit{t}-value & \textit{p}-value  \\
    \midrule
    (Intercept)  & 2.71845  &   0.61012  & 4.456 & <.001 *** \\
    Training data (TD) & 2.11411  &  0.55564  &  3.805 & \textbf{<.001} *** \\
    Irregularity (I) & 2.83197  &  0.42242  &  6.704 & \textbf{<.001} ***  \\
    TTR (T) &  -0.62728  &  0.57127 & -1.098 & 0.272179 \\
    TD*I & 2.03621  &  0.31730 &  6.417 & \textbf{<.001} *** \\
    TD*T & -0.62728  &  0.57127 & -1.098 & 0.272179 \\
    \midrule
    \textbf{Random Effect} \\
    \midrule
    Group & Name & Variance & Std.Dev. \\
    \midrule
    language & (Intercept) & 3.0920 &  1.7584 \\
    task & (Intercept)  & 0.3571 &  0.5976 \\
    \bottomrule
    \end{tabular}
\end{table*}
\end{document}